\documentclass[sigconf]{acmart}

\AtBeginDocument{%
  \providecommand\BibTeX{{%
    \normalfont B\kern-0.5em{\scshape i\kern-0.25em b}\kern-0.8em\TeX}}}

\setcopyright{acmcopyright}
\copyrightyear{2022}
\acmYear{2022}
\acmDOI{XXXXXXX.XXXXXXX}

\acmPrice{15.00}
\acmISBN{978-1-4503-XXXX-X/18/06}

\usepackage{subfigure}
\usepackage{multirow}
\usepackage{makecell}

\begin{document}

\title{VEM$^2$L: A Plug-and-play Framework for Fusing Text and Structure Knowledge on Sparse Knowledge Graph Completion}

\author{Tao He}
\affiliation{%
  \institution{Harbin Institute of Technology,}
  \country{China}
}
\email{the@ir.hit.edu.cn}

\author{Ming Liu}
\authornote{Corresponding Author.}
\affiliation{%
  \institution{Harbin Institute of Technology,}
  \country{China}
}
\email{mliu@ir.hit.edu.cn}

\author{Yixin Cao}
\affiliation{%
  \institution{Singapore Management University,}
  \country{Singapore}
}
\email{caoyixin2011@gmail.com}

\author{Tianwen Jiang}
\affiliation{%
  \institution{Harbin Institute of Technology,}
  \country{China}
}
\email{twjiang@ir.hit.edu.cn}

\author{Zihao Zheng}
\affiliation{%
  \institution{Harbin Institute of Technology,}
  \country{China}
}
\email{zhzheng@ir.hit.edu.cn}

\author{Jingrun Zhang}
\affiliation{%
 \institution{Harbin Institute of Technology,}
  \country{China}
}
\email{jrzhang@ir.hit.edu.cn}

\author{Sendong Zhao}
\affiliation{%
  \institution{Harbin Institute of Technology,}
  \country{China}}
\email{zhaosendong@gmail.com}

\author{Bing Qin}
\affiliation{%
  \institution{Harbin Institute of Technology,}
  \country{China}}
\email{qinb@ir.hit.edu.cn}


\begin{abstract}
  Knowledge Graph Completion (KGC) aims to reason over known facts and infer missing links but achieves weak performances on those sparse Knowledge Graphs (KGs). Recent works introduce text information as auxiliary features or apply graph densification to alleviate this challenge, but suffer from problems of ineffectively incorporating structure features and injecting noisy triples.
  In this paper, we solve the sparse KGC from these two motivations simultaneously and handle their respective drawbacks further, and propose a plug-and-play unified framework VEM$^2$L over sparse KGs.
  The basic idea of VEM$^2$L is to motivate a text-based KGC model and a structure-based KGC model to learn with each other to fuse respective knowledge into unity. To exploit text and structure features together in depth, we partition knowledge within models into two nonoverlapping parts: expressiveness ability on the training set and generalization ability upon unobserved queries. For the former, we motivate these two text-based and structure-based models to learn from each other on the training sets. And for the generalization ability, we propose a novel knowledge fusion strategy derived by the Variational EM (VEM) algorithm, during which we also apply a graph densification operation to alleviate the sparse graph problem further. 
  Our graph densification is derived by VEM algorithm.
  Due to the convergence of EM algorithm, we guarantee the increase of likelihood function theoretically with less being impacted by noisy injected triples heavily. 
  By combining these two fusion methods and graph densification, we propose the VEM$^2$L framework finally.
  Both detailed theoretical evidence, as well as qualitative experiments, demonstrates the effectiveness of our proposed framework. 
\end{abstract}






\maketitle

\section{Introduction}
Knowledge Graphs (KGs) have been widely used in many practical tasks, such as question answering \cite{huang2019knowledge}, information retrieval \cite{xiong2017explicit}, and recommendations \cite{wang2018dkn}. However, KGs are far from complete as inherent limitations of the technology and corpus. Knowledge Graph Completion (KGC) is a promising technique to solve this problem \cite{vashishth2019composition}. For a factual triple comprised of head entity, relation, and tail entity, KGC aims to automatically predict the missing one given two other elements. Previous methods mainly focus on modeling graph structure features within KGs \cite{rossi2021knowledge}, involving triple-level \cite{bordes2013translating, nickel2011three} and path-level structure features \cite{das2017go, chen2018variational, zhang2021learning}. 
However, structural features are often sensitive to the sparsity of KGs \cite{lv2020dynamic, fu2019collaborative}. As shown in Figure 1, there is a downward trend in KGC performance along with the increasing graph sparsity. That is, the sparser the KG is, the fewer neighbors exist, thus entities lack enough structure information to depict their relationships with others.

\begin{figure}[t]
    \centering
    \includegraphics[width=0.90\linewidth]{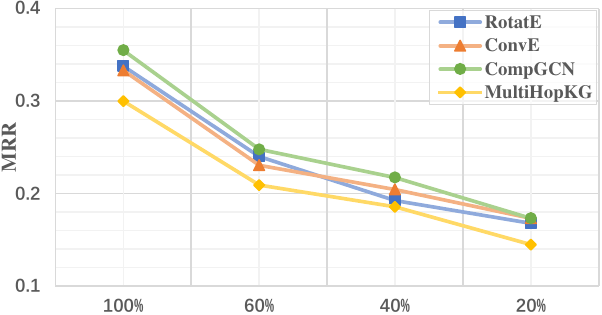}
    \caption{KGC MRR results of different structure-based models on FB15K-237 and its sparse subsets (60\%, 40\%, and 20\% denote percentages of retained triples).}
    \label{fig:conceptnet_mrr_curve}
\end{figure}

Relevant work addresses this challenge via two key motivations: 
(1) \textbf{Data augmentation, that is densifying KGs via generating new triples.} Lv et al. \cite{lv2020dynamic} magnify action spaces via adding new edges produced by KG Embedding models. 
Malaviya et al. \cite{malaviya2020commonsense} densify KGs using BERT \cite{devlin2018bert} to generate a series of undirected edges. 
However, it is inevitable that noisy triples are also injected during this process.
Noises of invalid triples may propagate over the graphs and cause harmful impacts on the learning processes. 
(2) \textbf{Auxiliary text information.} 
Entities and relations in existing KGs are usually accompanied with textual names or descriptions, which offer additional semantics for KG modeling \cite{xu2016knowledge, xiao2017ssp, an2018accurate}.
Recently, pre-trained language models (PLMs) \cite{devlin2018bert,liu2019roberta} have achieved great success and are able to obtain high-quality textual vectors. Although PLM-based KGC models have better generalization ability due to the massive implicit knowledge acquired from pre-training corpus, these methods are suspicious of ineffectively incorporating KG structures \cite{markowitz2022statik}.

In this paper, we unify these two motivations into one knowledge fusion framework for sparse KGC. 
Our framework consists of a structure-based and a text-based KGC model as knowledge extractors. After fully pre-training these two KGC models independently, we motivates them to exchange acquired knowledge, during which we also apply the text-based model to densify KGs to further relieve the graph sparsity.
Based on this preliminary idea, we target two key problems: \textbf{Q1} How to mutually enhance the two types of structural and textual features? \textbf{Q2} How to density KGs with the guarantee of relieving harmful impacts of noisy triples? 

For $\mathbf{Q1}$: We start from the thought of knowledge fusion to exploit structural and textual features together. Specifically, we partition knowledge within models into two non-overlapping parts: \textbf{expressiveness ability} reflected by the fitting degree on training data, and \textbf{generalization ability} reflected by the predictions on unobserved queries \cite{li2021does}. Correspondingly, we propose two different fusing methods for these two kinds of knowledge based on Mutual Learning (ML) \cite{zhang2018deep} and Variational EM (VEM) \cite{bishop2006pattern} respectively. The former method adopts Mutual Learning, an ensemble learning method that learns a group of models collaboratively, to motivate the structure-based and text-based models to learn from each other on training data. The latter applies VEM algorithm and motivates models to exchange generalization abilities over unobserved queries. After fully joint learning, the model that performs better is considered to have learned both text and structure knowledge. 

For \textbf{Q2}: 
For the VEM-based method, we start from the objective of maximizing the likelihood function and also derive the graph densification as the intermediate operation. 
We apply graph densification here as it is the necessary step for implementing the VEM-based method. Due to the convergence of EM algorithm, we can increase the likelihood function in theory with less being impacted by noisy injected triples heavily. 
On the other hand, PLMs are ideal implementation candidates for the text-based model as massive knowledge acquired from the pre-training corpus are stored in PLMs, and thus implicit triples generated by the PLM-based KGC model are expected to densify KGs and improve the generalization ability of the fusing model further.


By combining the ML-based and VEM-based fusion methods, we propose the VEM$^2$L framework finally. 
Our main contributions are highlighted as follows:
\begin{itemize}
    \item We solve the sparse KGC challenge from two aspects simultaneously: utilizing both structure and text information, and dynamically densifying the KGs.
    \item We propose a plug-and-play knowledge fusion framework VEM$^2$L consisting of ML-based and VEM-based methods. Different pre-trained text-based and structure-based models can be plugged into our framework compatibly.
    \item We propose a method of dynamical graph densification to alleviate sparse KGs further with the ability to relieve negative impacts of noisy injected triples.
    \item Through extensive experiments on three sparse benchmarks and detailed analyses of the importance of components and working mechanisms, we demonstrate the effectiveness of our proposed framework. 
\end{itemize}




\section{Related Work}
\noindent \textbf{Modeling Graph Structure.} Previous studies on KGC task mainly explore structural graph embedding approaches through spatial measurement or semantic matching in a low-dimension geometric space, such as TransE \cite{bordes2013translating}, DistMult \cite{yang2014embedding}, ComplEx \cite{trouillon2016complex}, ConvE \cite{dettmers2018convolutional}, and TuckER \cite{balavzevic2019tucker}.
There are also been effort in using graph neural networks (GNNs) \cite{kipf2016semi, velivckovic2017graph} for KGC task. R-GCN \cite{schlichtkrull2018modeling} modifies the original GCN for the multi-relational knowledge graph setting. Developed works like CompGCN \cite{vashishth2019composition}, KBGAT \cite{nathani2019learning}, and SACN \cite{shang2019end} proposed more powerful message aggregation functions and achieved promising performance. However, those previous methods just learn graph embeddings depending on the connective information between entities, which is sensitive to the graph sparsity.

\begin{figure*}[t]
    \centering
    \subfigure[Probabilistic Graphical Modeling]{
        \includegraphics[width=0.2\linewidth]{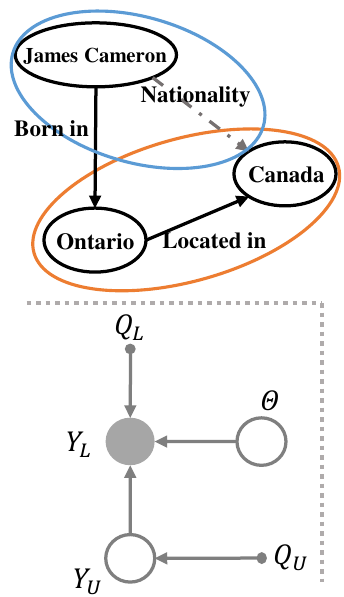}
    }
    \subfigure[Overview of VEM$^2$L Architecture]{
        \includegraphics[width=0.69\linewidth]{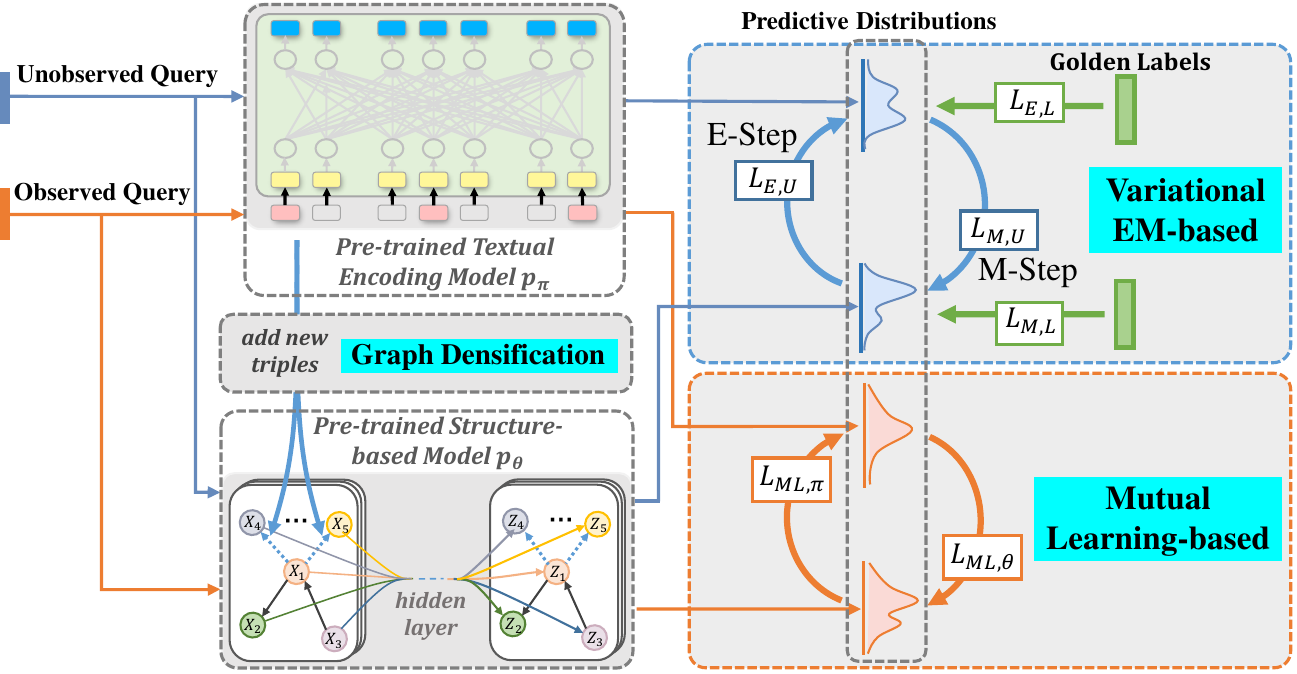}
    }
    \caption{Illustration of (a) the probabilistic graphic modeling for KGC task and (b) the VEM$^2$L framework, including VEM-based and ML-based parts. We use blue color to depict operations relevant to VEM-based method and orange color for ML-based. ML-based method trains each model in the training set with a supervised loss and a KL-based mimicry loss to match the distribution of their peers. VEM-based method also trains each model upon unobserved queries with mimicry losses after the graph densification operation. New triples are added to the KG generated by the text-based model to densify the graph.}
    \label{fig.main}
\end{figure*}

\noindent \textbf{Modeling Text Information.} Structure-based models have achieved great success while suffering from low graph sparsity. Apart from graph structure, textual information can be considered valuable supplementary for the KG representation learning task. 
DKRL \cite{xie2016representation} learns embeddings of entities from their descriptions using CNN. Xu et al. \cite{xu2016knowledge} encode valuable information from textual descriptions and integrate representations of structure and text with an attention mechanism. However, they are commonly less capable of exploiting textual information by only encoding descriptions without any prior knowledge.
Recent works have explored pre-trained language models to encode textual information. KG-BERT applies BERT \cite{devlin2018bert} to encode concatenated triples’ natural language text directly while such a simple method is criticized for terrible inefficiency and high overheads. 
Otherwise, the lack of graph structure information is also responsible for mediocre performances.
Instead of concatenating descriptions of the head entity and query relation, PKGC \cite{lv2022pre} converts each triple and its support information into natural prompt sentences. 
StTIK \cite{markowitz2022statik} utilizes both the structure as well as the underlying textual descriptions of entities while focusing on the inductive setting.
Other work like StAR \cite{wang2021structure} attempts to augment text representation with structural constraints by designing structure-augmented scores. While this work prefers more on text information but less on the graph structure, which we claim to be insufficient. By comparison, we fuse knowledge into unity by balancing the weight of text and graph structure dynamically.



\section{Notations and Problem Formulation}
A KG is represented by $\mathcal{G}=(\mathcal{E},\mathcal{R},\mathcal{T},\mathcal{D})$, where $\mathcal{E}$, $\mathcal{R}$ denote the entity and relation set, $\mathcal{T}$ is the set of all factual triples within $\mathcal{G}$, involving explicit facts $\mathcal{T}_L$ (i.e. all the triples that we know currently) and implicit facts $\mathcal{T}_U$ (i.e. triples that we do not know but indeed valid), and $\mathcal{D}=\{d^{(e_1)},...,d^{(e_{|E|})},d^{(r_1)},...,d^{(r_{|R|})}\}$ represents the set of names (or descriptions) for all entities and relations.
Each triple $(h,r,t)$ consists of a query $q=(h,r,?)$ and a label $t$, thus triple set $\mathcal{T}$ can be divided into a query set $Q=\{q_1,...,q_N\}$ and a label set ${Y}=\{y_1,...,y_N\}$, where $y_m=\{e|(h,r,e) \textit{ is valid, } e\in \mathcal{E}\}$ represents true labels for query $q_m=(h,r,?)$. Correspondingly, we also devide $\mathcal{T}_L$ (or $\mathcal{T}_U$) into $Q_L$ (or $Q_U$) and ${Y}_L$ (or  ${Y}_U$) for the same reason. 
Here, we will define some concepts for the sake of the following exposition.

\noindent{\textbf{Definition 1}} (Knowledge Graph Completion)
KGC task aims to predict $t$ given $(h,r,?)$ or $h$ given $(?,r,t)$, and without loss of generality, we denote both directions as query $(h,r,?)$ by adding reverse relations for query $(?,r,t)$. Under such denotation, we predict tail entities by calculating probability $pr(y_m|q_m)$. More generally, we train KGC models by maximizing $pr({Y}_L|Q_L)$ and predict unseen queries by calculating $pr({Y}_U|Q_U, \mathcal{T}_L)$.

\noindent{\textbf{Definition 2}} (Triple's Neighbors)
For a certain triple $t_n$, we define those triples interconnected with it as neighbor triples of $t_n$. For example, $(e_1,r_1,e_2)$ is a neighbor of $(e_3,r_2,e_1)$ because $e_1$ is a common entity for these two triples.

\section{Our Framework: VEM$^2$L}
In this section, we introduce our fusion framework VEM$^2$L to improve the performance of the KGC task over sparse KGs, which is comprised of a Variational EM-based (shorted as \textbf{VEM}-based) method and a Mutual Learning-based (abridged as \textbf{M}L-based) method. Figure 2(b) provides an overview of our framework. 

Firstly, we pre-train two text-based and structure-based KGC models until convergence to fully learn text and structure knowledge. Then our framework re-trains these two models jointly to fuse knowledge they have acquired. The ML-based method receives observed queries and motivates these two models to learn from each other via minimizing the KL divergences of output distributions on the input observed queries. On the other hand, the VEM-based method receives unobserved queries and encourages two models to exchange generalization capacity via minimizing the KL divergences of output distributions similarly on unobserved queries. The VEM-based method first applies E-step learning and then M-step. During the E-step learning, we densify the KGs with extra neighbor triples for the input unobserved queries produced by the text-based model, which we call it graph densification operation.
By combining ML-based and VEM-based methods, we propose a novel fusion framework VEM$^2$L for the sparse KGC task. After fully joint learning, the model that performs better is considered to have learned both text and structure knowledge.

We first introduce the ML-based fusing method, before presenting the VEM-based method. 
Finally, we re-train the VEM-based and ML-based methods collaboratively and introduce the overall learning process of the unified VEM$^2$L framework. 



\subsection{Mutual Learning-based Fusion}

Mutual Learning (ML) \cite{zhang2018deep} can be considered as a specific method of Knowledge Distillation \cite{hinton2015distilling, jafari2021annealing}.
The theory of ML is that models initialized with diverse start-points might search different paths to the optimum, thus making models less likely to be lost in locally optimal solutions. The learning process can be described as follows:
\begin{enumerate}
    \item For the same batch of training cases, network $\Theta_1$ and $\Theta_2$ predict their own distributions $pr_1$ and $pr_2$;
    \item Update $\Theta_1$ by minimizing the summation between $\Theta_1$'s cross-entropy loss $\mathcal{L}_1$ relative to golden labels and KL divergence $D_{KL}(\mathbb{SG}(pr_2)||pr_1)$, where $\mathbb{SG}$ refers to ``Stop Gradient''. ;
    \item Re-calculate $pr_1$ and update $\Theta_2$ by minimizing the summation between $\Theta_2$'s cross-entropy loss $\mathcal{L}_2$ relative to golden labels and KL divergence $D_{KL}(\mathbb{SG}(pr_1)||pr_2)$.
\end{enumerate}

In this paper, we adopt the key idea of ML to transfer knowledge between the structure-based model $p_\theta$ and text-based model $p_\pi$, where $\theta$ and $\pi$ are parameters of these two models respectively. To avoid redundancy, we only introduce mimicry losses here and mention golden label losses in the VEM-based part. Given an observed query $q$ with a golden label set $y \in \mathcal{Q_L}$, firstly, we compute prediction distributions $p_\theta(y|q)$ and $p_\pi(y|q)$ for models $p_\theta$ and $p_\pi$. Then we motivate $p_\pi$ to learn from $p_\theta$ by minimizing:
\begin{equation}
    \mathcal{L}_{ML,\pi} = D_{KL}(\mathbb{SG}(p_\theta(y|q))||p_\pi(y|q))
\end{equation}
After that, we re-compute $p_\pi(y|q)$ and drive $p_\theta$ to learn from $p_\pi$ by minimizing:
\begin{equation}
    \mathcal{L}_{ML,\theta} = D_{KL}(\mathbb{SG}(p_\pi(y|q))||p_\theta(y|q))
\end{equation}

Evidently, the above fusion method processes on the training set, thus models can exchange knowledge about how to fit observed cases, i.e. fitting capacity, with each other.

Here we want to emphasize the difference between our method and the original ML. Original ML trains students on the same data, thus knowledge difference is attributed to the capacity difference of models. While we fuse knowledge on different data, i.e. the graph structure and text information for KGs, which store diverse knowledge separately. Therefore, apart from knowledge relevant to model learning abilities, knowledge differences due to the data formats can also be fused in our method.

\subsection{Variational EM-based Fusion}

\subsubsection{Probabilistic Graphical Modeling}
Different from independent identically distributed hypothesis, i.e. i.d.d., graph data, especially for KGs, organizes graph structure through interconnected nodes. Under this observation, triples are tenable via supporting each other. When models are trained, the reasoning process for the observed query can be easier if useful implicit evidence exists. Figure 2(b) provides an intuitive explanation. For example, fact \textit{(James Cameron, nationality, Canada)} is unobserved but is valid, thus we call it implicit. During the training phase, models are trained to be able to reason the explicit fact \textit{(Ontario, part of, Canada)}. Obviously could this reasoning process become more straightforward if we have seen the fact of \textit{(James Cameron, nationality, Canada)} in advance. Consequently, we model the implicit triples as hidden variables for training cases, that is we consider calculating $pr(Y_L|{Q_L}, Q_U)$ rather than $pr({Y_L}|{Q_L})$. The probabilistic graph is constructed in Figure 2(a).

Based on this modeling result, we start to train structural model $p_\theta$ by maximizing the log-likelihood on observed facts, that is:
\begin{equation}
    \max_\theta \log p_\theta({Y_L|Q_L,Q_U};\theta)
\end{equation}
We further add unobserved triples as hidden factors for reasoning training cases, and partition the log-likelihood function as the summation of the evidence lower bound (ELBO) and KL-divergence:
\begin{equation}
    \begin{aligned}
        \log p_\theta(Y_L|Q_L, Q_U)&=\mathbb{E}_{p_\pi(Y_U|Q_U)}[\log\frac{p_\theta(Y_L,Y_U|Q)}{p_\pi(Y_U|Q_U)}]\\
        & \quad+D_{KL}(p_\pi(Y_U|{Q_U})||p_\theta({Y_U}|{Y_L, Q}))\\
        &=\mathbb{ELBO}+D_{KL}(p_\pi({Y_U}|{Q_U})||p_\theta({Y_U}|{Y_L}, Q))
    \end{aligned}
\end{equation}
where $p_\pi(Y_U|T_U)$ serves as the variational distribution implemented by the text-based model $p_\pi$. Next, we apply EM algorithm to maximize this log-likelihood function \cite{neal1998view}.

\subsubsection{E Step: Inference}
E-step aims to estimate ELBO by computing the posterior distribution $p_\theta(Y_U|Q, Y_L)$. However, exact inference is intractable due to the complicated relational structures. Therefore, Variational EM algorithm approximates $p_\theta(Y_U|Q, Y_L)$ with another variational distribution $p_\pi(Y_U|Q_U)$. \textbf{To fuse structure knowledge along with text knowledge, we apply a textual encoding model to implement $p_\pi$. So we can ignore the complicated graph structure and apply the mean-field hypothesis to factorize $p_\pi(Y_U|Q_U)$:}
\begin{equation}
    p_\pi(Y_U|Q_U)\approx \prod_{q_m\in Q_U}p_\pi(y_m|q_m)
\end{equation}
where we denote $p_\pi(y_m|q_m)$ as the probability that $y_m$ is the label of query $q_m$ provided by the text-based model $p_\pi$.

Next, $p_\pi(Y_U|Q_U)$ is approximated by minimizing the KL divergence in Eq.(4). According to the vanilla variational inference algorithm, we introduce Theorem 1 to show the fixed-point condition that $p_\pi(Y_U|Q_U)$ needs to satisfy.

\noindent{\textbf{Theorem 1}} Under the mean filed hypothesis, minimizing the KL divergence $D_{KL}(p_\pi({Y_U}|{Q_U})||p_\theta({Y_U}|{Y_L}, Q))$ is equivalent to optimize the following objective:
\begin{equation}
    \begin{aligned}
    &\max_\pi \quad \sum_{t_m \in \mathcal{T}_U}
    \{ \mathcal{H}(p_\pi(y_m|q_m)) \\ &+\sum_{y_m}\mathbb{E}_{p_\pi(Y_{\mathrm{NB}(t_m)\cap \mathcal{T}_U})}[\log p_\theta(y_m|q_m,\mathcal{T}_{\mathrm{NB}(t_m)})] \}
    \end{aligned}
\end{equation}
where we denote $p_\pi(Y_{\mathrm{NB}(t_m)\cap \mathcal{T}_U}|Q_{\mathrm{NB}(t_m)\cap \mathcal{T}_U})$ as $p_\pi(Y_{\mathrm{NB}(t_m)\cap \mathcal{T}_U})$ for brevity. $\mathrm{NB}(t_m)$ is the neighbor set of triple $t_m$, thus $\mathrm{NB}(t_m) \cap \mathcal{T}_U$ represents the intersection of $\mathrm{NB}(t_m)$ and unobserved triples set $\mathcal{T}_U$. We also use $p_\theta(y_m|q_m,\mathcal{T}_{\mathrm{NB}(t_m)})$ to represent the probability that $y_m$ is the label of query $q_m$ given the neighbor triple set $\mathcal{T}_{\mathrm{NB}(t_m)}$, computed by structure-based model $p_\theta$. Here, $\mathcal{H}(p_\pi(y_m|q_m))$ denotes the entropy value of the distribution $p_\pi(y_m|q_m)$. 
The proof process is guided by vanilla Variational Inference \cite{bishop2006pattern}. And we present it in Appendix A.

To further simplify Eq.(6), we estimate the expectation formula by sampling from $p_\pi(Y_{\mathrm{NB}(t_m)\cap \mathcal{T}_U})$, resulting in:
\begin{equation}
    \begin{aligned}
        &\quad \mathbb{E}_{p_\pi(Y_{\mathrm{NB}(t_m)\cap T_U})}[\log p_\theta(y_m|q_m, Q_{\mathrm{NB}(t_m)}, Y_{\mathrm{NB}(t_m)})]\\
        &\approx \log p_\theta(y_m|q_m, Q_{\mathrm{NB}(t_m)}, \widehat{Y}_{\mathrm{NB}(t_m)})
    \end{aligned}
\end{equation}
where $Q_{\mathrm{NB}(t_m)}$ denotes the query set of neighbor triple set $\mathrm{NB}(t_m)$. $\widehat{Y}_{\mathrm{NB}(t_m)}$ is sampled labels for neighbor triples of $t_m$. Specifically, for each neighbor triple, we use the ground-truth label if it is observed, otherwise we sample its label from $p_\pi(Y_{\mathrm{NB}(t_m)\cap \mathcal{T}_U})$. The sampling method for unlabeled neighbors is introduced in algorithmic details following. During this process, a certain number of neighbor triples are added into the KG, thus we call this process \textbf{graph densification}, by which the number of neighbors for the head entity will increase and thus graph sparseness can be further alleviated.

By estimating Eq.(6) via Eq.(7), we get the objective of E-step as:
\begin{equation}
    \mathcal{L}_{E,U}=\sum_{t_m\in \mathcal{T}_U}D_{KL}(p_\pi(y_m|q_m)||\mathbb{SG}(p_\theta(y_m|q_m,\widehat{Y}_{\mathrm{NB}(t_m)})))
\end{equation}
In E-step, we fix $p_\theta$ and optimize $p_\pi$ to minimize $\mathcal{L}_{E,U}$. The practical idea of this objective is clear: $p_\pi$, i.e. textual encoder, is supposed to learn from $p_\theta$, i.e. structural encoder, over unobserved queries after the graph densification process, thus generalization ability learnt from structure messages is flowed into the textual encoder.

To avoid the semantic drift, $p_\pi$ is also trained on training cases as additional loss function:
\begin{equation}
    \mathcal{L}_{E,L}=-\sum_{t_n\in \mathcal{T}_L}\log p_\pi(\hat{y}_n|q_n)
\end{equation}
where $\hat{y}_n$ is true label set for observed query $q_n$. Thus we optimize $p_\pi$ by minimizing $\mathcal{L}_{E}=\mathcal{L}_{E,U} + \mathcal{L}_{E,L}$ finally.

\subsubsection{M Step: Learning} 
M-step aims to maximize the expectation formula by learning $p_\theta$ with $p_\pi$ fixed, resulting in:
\begin{equation}
    \mathbb{E}_{p_\pi(Y_U|Q_U)}[\log p_\theta(Y_L,Y_U|Q)]
\end{equation}
It is difficult to maximize Eq.(10) directly for the existence of partition function. Therefore, we change to optimize the pseudolikelihood function \cite{besag1975statistical}. Here, we introduce Theorem 2 to provide an approximate conclusion for Eq.(10).

\noindent{\textbf{Theorem 2}} Maximizing $\mathbb{E}_{p_\pi(Y_U|Q_U)}[\log p_\theta(Y_L,Y_U|Q)]$ is equivalent to minimizing the following objective:
\begin{equation}
    \begin{aligned}
    \mathcal{L}_{M}=&-\sum_{t_n\in \mathcal{T}_U}\mathbb{E}_{p_\pi(y_n|q_n)}[\log p_\theta(y_n|q_n, Y_{\mathrm{NB}(t_n)})]\\
    &-\sum_{t_m\in \mathcal{T}_L}\log p_\theta(\hat{y}_m|q_m, Y_{\mathrm{NB}(t_m) \cap \mathcal{T}_L})\\
    =&\mathcal{L}_{M,U}+\mathcal{L}_{M,L}
    \end{aligned}
\end{equation}
where $\mathrm{NB}(t_m) \cap \mathcal{T}_L$ represents the intersection of $t_m$'s neighbor triples set $\mathrm{NB}(t_m)$ and training triples set $\mathcal{T}_L$. The process of proof is motivated by GMNN \cite{qu2019gmnn}. And we present it in Appendix A.

Intuitively, $\mathcal{L}_M$ consists of two parts: $\mathcal{L}_{M,U}$ requires $p_\theta$ to keep pace with $p_\pi$ for the same unobserved query, thus $p_\theta$ acquires complementary generalization knowledge from $p_\pi$; $\mathcal{L}_{M,L}$ requires $p_\theta$ to continue to receive the guidance of golden labels to avoid semantic drift same as in E-Step.

\subsubsection{Overview} In E-step, the textual model $p_\pi$ learns generalization capacity from the structure-based model $p_\theta$ after graph densification. 
After that, M-step optimizes $p_\theta$ to learn from textual model $p_\pi$ upon unobserved queries, thus generalization ability learned from the text also flows into the structure-based model.

Knowledge about generalization ability is transferred between structure-based and text-based models iteratively, \textbf{during which the log-likelihood function keeps converging as the convergence property of EM algorithm, thus the fusion framework can guarantee the improvement theoretically even if there are invalid edges being injected during graph densification.}

\subsection{VEM$^2$L Fusion Framework Learning}
On the whole, we introduce the learning process of our fusion framework in this section.

The overall fusion framework iterates between two training phases: (1) Motivating the text-based model to learn from the structure-based one. (2) Driving the structure-based model to learn from the text-based one. Specifically, given two pre-trained KGC models $p_\theta$ and $p_\pi$, VEM$^2$L first applies the text-based model $p_\pi$ to enrich the neighbors of constructed queries. Hereafter, VEM$^2$L fixes structure-based model $p_\theta$ and optimizes text-based model $p_\pi$ by minimizing $\mathcal{L}_{E,L} + \alpha_t\mathcal{L}_{E,U} + \alpha_s\mathcal{L}_{ML,\pi}$. And then $\mathcal{L}_{M,L} + \beta_t\mathcal{L}_{M,U} + \beta_s\mathcal{L}_{ML,\theta}$ is optimized on the original KGs by updating $p_\theta$ with $p_\pi$ fixed.
This learning process will be repeated alternately for certain epochs. 

\section{Implementation Details}
In this section, we introduce some implementation details for better understanding.

\subsection{Implementations for Textual Encoding Model and Structure-based Model}
\noindent{\textbf{Textual Encoding Model.}} We follow Yao et al. \cite{yao2019kg} and model $p_\pi$ with BERT for its excellent text encoding performance. Different from KG-BERT \cite{yao2019kg}, we input a query $(h,r,?)$ and predict true tails according to the output of $[CLS]$ position, instead of judging the correctness of triples directly. In this way, we have greatly relieved the heavy overheads problem of KG-BERT. Given a query $(h,r,?)$, we concatenate the text of head entity $h$ and relation $r$ as follows:
\begin{equation}
    [x_{[CLS]}, d^{(h)}, x_{[SEP]}, d^{(r)}, x_{[SEP]}]
\end{equation}
Then the concatenated sequential tokens are fed into BERT(base) encoder. Finally, outputs of $[CLS]$ in the last layer serves as inputs for the final classifier. We call this model as tinyKG-BERT.

\begin{table*}[t]
    \centering
    \caption{Experimental results on ConceptNet-100K, WN18RR and FB15K-237\_20 test sets. *Results of running MultiHopKG\cite{lin2018multi} and DacKGR\cite{lv2020dynamic} on both head and tail prediction settings. Hits@N and MRR values are in percentage. The best score is in \textbf{bold}.}
    \begin{tabular}{cccccccccccccccc}
    \toprule
    &&&& \multicolumn{4}{c}{CN-100K} & \multicolumn{4}{c}{WN18RR} & \multicolumn{4}{c}{FB15k-237\_{20}}\\
    \cmidrule(lr){5-8}\cmidrule(lr){9-12}\cmidrule(lr){13-16}
    &&&& \multicolumn{3}{c}{Hits@N $\uparrow$} && \multicolumn{3}{c}{Hits@N$\uparrow$} && \multicolumn{3}{c}{Hits@N$\uparrow$}\\
    \cmidrule(lr){5-7}\cmidrule(lr){9-11}\cmidrule(lr){13-15}
    &&&& @1 & @3 & @10 & MRR $\uparrow$ & @1 & @3 & @10 & MRR $\uparrow$ & @1 & @3 & @10 & MRR $\uparrow$\\
    \midrule
    \multicolumn{4}{l}{TransE \cite{bordes2013translating}} & 6.04 & 30.71 & 50.71 & 21.56 & 4.3 & 44.1 & 53.2 & 24.3 & 9.84 & - & 29.93 & 16.54\\
    \multicolumn{4}{l}{RotatE \cite{sun2019rotate}} & 16.21 & 35.29 & 53.17 & 28.76 & 42.8 & \underline{49.2} & 57.1 & 47.6 & 10.15 & 17.89 & 30.33 & 16.80\\
    \multicolumn{4}{l}{ComplEx \cite{trouillon2016complex}} & 19.33 & 30.75 & 44.33 & 27.62 & 41.00 & 46.00 & 51.00 & 44.00 & 6.03 & 11.51 & 21.06 & 11.03\\
    \multicolumn{4}{l}{TuckER \cite{balavzevic2019tucker}} & \underline{20.33} & 30.75 & 45.13 & 28.29 & \underline{44.30} & 48.20 & 52.60 & \underline{47.00} & 11.81 & 18.75 & 30.24 & 17.89\\
    \multicolumn{4}{l}{InteractE \cite{vashishth2020interacte}} & 19.71 & 32.96 & 46.46 & 28.97 & 43.00 & - & 52.80 & 46.30 & \underline{12.02} & \underline{19.33} & 30.70 & \underline{18.27}\\
    \multicolumn{4}{l}{SACN \cite{shang2019end}} & - & - & - & - & 43.00 & 48.00 & 54.00 & \underline{47.00} & 11.08 & 17.83 & 28.61 & 16.93\\
    \multicolumn{4}{l}{MultiHop \cite{lin2018multi}*} & - & - & - & - & 39.28 & 43.30 & 48.85 & 42.46 & 9.67 & 15.62 & 24.50 & 14.50\\
    \multicolumn{4}{l}{DacKGR \cite{lv2020dynamic}*} & - & - & - & - & 27.43 & 37.03 & 45.15 & 33.58 & 10.44 & 16.47 & 24.92 & 15.12\\
    \multicolumn{4}{l}{KG-BERT \cite{yao2019kg}} & - & - & - & - & 4.1 & 30.2 & 52.4 & 21.6 & - & - & - & -\\
    \multicolumn{4}{l}{StAR \cite{wang2021structure}} & 16.71 & \underline{34.42} & \underline{56.67} & \underline{29.59} & 24.3 & 49.1 & \underline{\textbf{70.9}} & 40.1 & 10.18 & 17.88 & \underline{30.95} & 17.07\\
    \midrule
    \multicolumn{4}{l}{CompGCN \cite{vashishth2019composition}} & 21.08 & 32.96 & 47.96 & 30.00 & 43.28 & 48.05 & 53.14 & 46.74 & 11.38 & 18.21 & 29.47 & 17.34\\
    \multicolumn{4}{l}{tinyKG-BERT} & 25.25 & 42.38 & 58.83 & 36.67 & 50.53 & 57.72 &  64.68 & 55.51 & 11.80 & 18.11 & 28.63 & 17.38\\
    \midrule
    \multicolumn{4}{l}{VEM-based} & 27.17 & 45.00 & 60.75 & 38.79 & 51.04 & 57.98 & 65.24 & 55.95 & 12.15 & 19.53 & 31.27 & 18.52\\
    \multicolumn{4}{l}{ML-based} & 27.54 & 45.67 & 62.5 & 39.49 & \underline{\textbf{51.95}} & 58.09 & 65.19 & \underline{\textbf{56.53}} & 12.29 & 20.14 & 32.13 & 18.90\\
    \multicolumn{4}{l}{VEM$^2$L} & \underline{\textbf{30.67}} & \underline{\textbf{48.53}} & \underline{\textbf{65.21}} & \underline{\textbf{42.41}} & 51.56 & \underline{\textbf{58.26}} & \underline{{65.78}} & 56.38 & \underline{\textbf{12.53}} & \underline{\textbf{20.30}} & \underline{\textbf{32.25}} & \underline{\textbf{19.12}}\\
    \bottomrule
    \end{tabular}
    \label{tab:table_1}
\end{table*}

\noindent{\textbf{Structure-based Model.}}
We make no more assumptions for $p_\theta$ except Markov Independence Hypothesis. For efficiency and effectiveness, we choose CompGCN to implement $p_\theta$ network. Of course, our framework is agnostic to the particular choice of GNN-based KGC model, as long as it follows the message aggregation mechanism.

\subsection{Graph Densification}
For VEM-based method, we need to generate neighbor triples for unobserved queried triples to estimate expectation in Eq.(7), which we call the graph densification process. By this operation, more possible triples are generated to increase neighbors of queries, playing a positive effect on relieving the graph-sparsity problem. 
Specifically, the graph densification process is divided into two sub-steps: (1) we first generate unobserved triple $t_m\in \mathcal{T}_U$ as the query; (2) We sample unlabeled neighbor triples for $t_m$ from $p_\pi(Y_{NB(t_m)\cap \mathcal{T}_U})$. 

\subsubsection{Generating Unobserved Triple $t_m$} 
Firstly, we sample an entity $e \in \mathcal{E}$ randomly. 
For each head entity $e$ in the training mini-batch, we obtain a relation set $\mathcal{R}_1$ in which each relation is connected to $e$, and the other unconnected relation set $\mathcal{R}_2 = \mathcal{R} \backslash \mathcal{R}_1$. Then we choose a query relation $r$ from $\mathcal{R}_2$ according to the embedding cosine similarity score between all pairs of relations in $\mathcal{R}_1$ and $\mathcal{R}_2$. Evidently, triple $(e,r,?)$ can not be found in the training set regardless of $``?"$ is. Of course, the generation method mentioned above is too simple to ensure enough correct triples but is low-complexity. And any other generation methods could be applied here. The relation embedding is provided by the structure-based model. 
This process will be repeated $\mathcal{N}$ times to generate $\mathcal{N}$ unlabeled $t_m$. 

\subsubsection{Sampling Unobserved Neighbor Triples}
According to Eq.(7), we need to estimate the expectation by sampling neighbors from $p_\pi(Y_{\mathrm{NB}(t_m)\cap \mathcal{T}_U})$. However, we have no idea about $\mathcal{T}_U$, let alone $\mathrm{NB}(t_m) \cap \mathcal{T}_U$. Therefore, we apply the same strategy stated above to generate $\mathcal{M}$ unlabeled neighbor triples. After that, we use $p_\pi$ to predict labels for them thus we succeed in obtaining $\mathcal{M}$ unobserved neighbor triples for $t_m$. After that, we import these $\mathcal{M}$ sampled neighbor triples into Eq.(7) to estimate the expectation.

\section{Experiments}
In this section, 
we introduce benchmarks and setups applied in our experiments and evaluate the effectiveness of our framework. To analyze the importance of the various components and the working mechanism for VEM$^2$L, we present the results of some detailed ablation and case studies.


\subsection{Experimental Settings}

\subsubsection{Datasets} We apply three prevalent KGC benchmarks: FB15k-237, WN18RR, and ConceptNet-100K (CN-100K) for evaluation. WN18RR \cite{dettmers2018convolutional} and FB15k-237 \cite{toutanova2015representing} are extracted from WN18 and FB15k \cite{bordes2013translating} respectively by removing data leakage. CN-100K \cite{li2016commonsense} is a sparse commonsense KG and contains the Open Mind Common Sense (OMCS) entries from ConceptNet \cite{speer2013conceptnet}. However, we observe that FB15k-237 is much denser than others.
To explore the performance of our framework within sparse scenarios, we uniformly extract 20\% number of triples from FB15k-237 to construct a sparser dataset, abridged as FB15k-237\_20. Differ from previous work \cite{lv2020dynamic}, we keep the contained entities and relations unchanged by constraining each entity (relation) to participate in at least one triple fact. The sparser KGs than FB15k-237\_20 would contain plenty of triples existing without triple neighbors, which is unfair to path-based baselines \cite{lin2018multi, lv2020dynamic}. In line with previous approaches, we employ descriptions and names as text information for WN18RR and CN-100K respectively. For FB15k-237\_20, we apply names for entities and relations directly for the reason of sequence length limits. The statistical details are summarized in Table 2.

\begin{table}[t]
    \centering
    \caption{Summary statistics of datasets}
    \begin{tabular}{l|cccc}
    \toprule
     & FB15k-237\_20 & WN18RR & CN-100K \\
    \midrule
    \#Entity & 14541 & 40943 & 78334\\
    \#Relation & 237 & 11 & 34 \\
    \#Training Triples & 54423 & 86835 & 10k \\
    \#Dev Triples & 17535 & 3034 & 1.2k \\
    \#Test Triples & 20466 & 3134 & 1.2k \\
    Average Out Degree & 3.752 & 2.141 & 1.731 \\
    MaxLen (Entity Desc.) & 1009 & 147 & 35 \\
    \bottomrule
    \end{tabular}
    \label{tab:table_2}
\end{table}

\begin{figure}[t]
    \centering
    \includegraphics[width=0.93\linewidth]{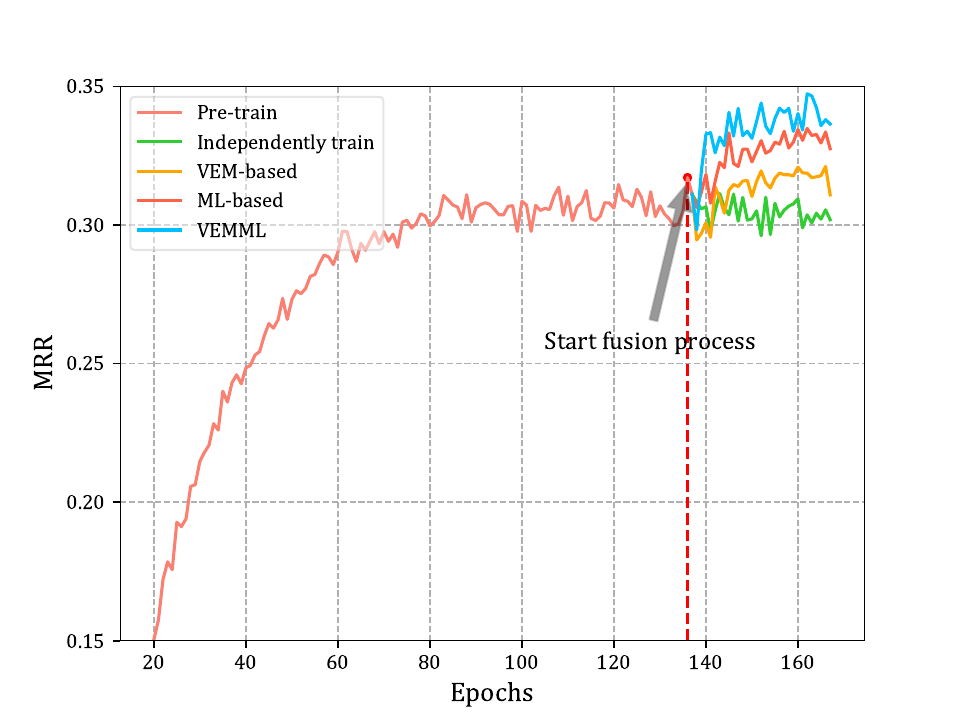}
    \caption{Convergence curves on CN-100K. We present numbers of epochs (x-axis) versus validation MRR (y-axis). For better visualization, we cut the range of epochs to [20, 170].}
    \label{fig:conceptnet_mrr_curve}
\end{figure}

\subsubsection{Evaluation Metrics} Same as previous work, we use ranking metrics to evaluate our framework, i.e. MRR and HITS@k. Besides, we filter out all remaining entities valid for the test query $(h, r, ?)$ from the ranking. The metrics are measured in both directions, where we calculate the ranking of $t$ for query $(h, r, ?)$ and the ranking of $h$ given $(t, r^{-1}, ?)$, where $r^{-1}$ refers to the inverse relation of $r$. Thus we report the mean MRR and HITS scores averaged across both directions. We strictly follow the “RANDOM” protocol proposed by Sun et al. \cite{sun2019re} to evaluate our methods.

\subsubsection{Baselines}
We conduct fusion experiments using CompGCN \cite{vashishth2019composition} and BERT \cite{devlin2018bert}, thus we compare the effectiveness before and after the joint learning process to display the relative increase. Besides, we compare VEM$^2$L with representative structure-based models \cite{bordes2013translating, sun2019rotate, trouillon2016complex, balavzevic2019tucker, vashishth2020interacte, shang2019end}, path-based model \cite{lin2018multi} and its improved version for sparse KGC \cite{lv2020dynamic}, text-based method KG-BERT \cite{yao2019kg}, and the state-of-the-art joint learning work StAR \cite{wang2021structure}.

\subsubsection{Training Regimen}
Wee first pre-train two text-based and structure-based models independently until convergence. Hereafter, we jointly train them for certain epochs to fuse knowledge acquired during pre-training and select the superior one on the valid set as the evaluation model. Specifically, we select tinyKG-BERT for WN18RR and CN-100K as the evaluation model and CompGCN for FB15k-237\_20 in practice. 
For ablation, we fuse models by the means of VEM-based and ML-based methods respectively to verify their effectiveness alone.

\subsubsection{Hyperparameters}
Based on the best MRR on the dev set, we set learning rates 0.008/0.00005 for CompGCN/tinyKG-BERT on FB15k-237\_20, 0.0009/0.0008 for WN18RR, and 0.00008/0.0005 for CN-100K. We set the training batch size to 64, embedding dimension to 128 for CompGCN on all benchmarks. Above hyper-parameters are retained unchangeable before and after fusion processes. For fusion experiments, the number of generated queries $\mathcal{N} = 32$ after grid searching within \{16, 32, 64\}, and $\mathcal{M} = 8$ after grid searching within \{1, 4, 8, 16\}. We jointly train 30 epochs totally for FB15K-237\_20 and CN-100K, and 20 epochs for WN18RR.
More parameter settings are presented in Appendix B.

\subsection{Link Prediction Results}
Table 3 shows the overall evaluation result of VEM$^2$L against baseline methods. 

As expected, the VEM$^2$L framework achieves improvements on all datasets. Specifically, the improvement is especially prominent on CN-100K, which achieves +6.375 points and +5.743 points improvements at Hits@10 and MRR respectively compared with learning dependently. The performance on FB15k-237\_20 achieves +2.78 points and +1.78 points improvements at Hits@10 and MRR.
To further validate the stability and effectiveness of our framework, we quantitatively illustrate how MRR changes as the training process goes in Figure 3. There is a clear trend that the performance of VEM$^2$L improves rapidly during the fusion process.
We find that the improvements are mainly contributed by combining two fusion methods via the ablation study mentioned below.

Weak improvements for WN18RR can be attributed to the noisy neighbor triples sampled in the graph densification operator, for which we generate unobserved queries via similarities between relation embeddings. However, we find semantics of relations on the WN18RR vary considerably from each other. And the fewer semantically similar relations exist, the more noisy triples might be sampled. To illustrate this point, we present heatmaps for relation similarities for WN18RR and CN-100K in Figure 4. Relation similarity is computed by counting Jaccard similarity for connected head entities set. Here we hypothesize that two relations shared with more head entities are more similar. We find the semantic differences of relations on the WN18RR are larger than those on the CN-100K. Improvements fail to live up to our expectations on the WN18RR, but it still remarkably outperforms other baselines in the terms of MRR and Hits@1.
Besides, worse performances of DacKGR \cite{lv2020dynamic} than MultiHop \cite{lin2018multi} on WN18RR confirm the negative impacts of noisy injected triples mentioned in the introduction section. 

\begin{table}[t]
    \centering
    \caption{Improvements of VEM$^2$L with different $\mathcal{M}$ on CN-100K. Values are in percentage.}
    \begin{tabular}{c|ccccc}
    \toprule
     & $\mathcal{M}$\textbf{=0} & \textbf{=1} & \textbf{=4} & \textbf{=8} & \textbf{=16} \\
    \midrule
    MRR & +4.92 & +4.822 & +5.242 & +5.743 & +4.651 \\
    Hits@1 & +4.625 & +4.875 & +5.292 & +5.417 & +4.417 \\
    \bottomrule
    \end{tabular}
    \label{tab:table_2}
\end{table}

\begin{figure}[t]
    \centering
    \subfigure[WN18RR]{
	    \includegraphics[width=1.6in]{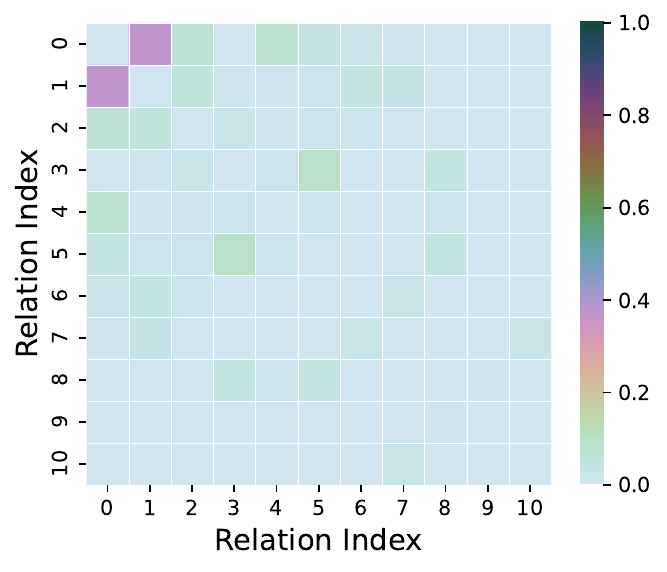}
    }
    \subfigure[CN-100K]{
        \includegraphics[width=1.5in]{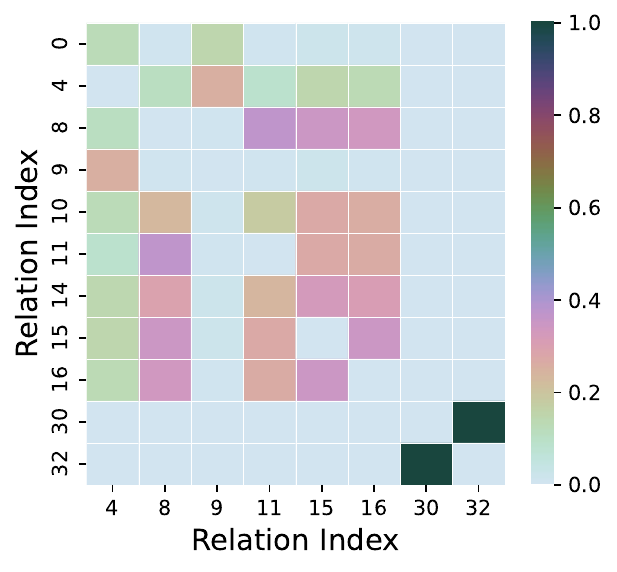}
    }
    \caption{Relation similarities between different pairs of relations. For CN-100K, we choose 11 relations with top-11 similarity values from all 34 relations for comparison.}
    \label{relation similarity}
\end{figure}

\begin{table*}[t]
    \centering
    \caption{Top-5 ranking results of candidate entities in CN-100K test. The first column includes queries for inference, and the second column presents their gold answers. The others include the ranking position and Top-5 ranked candidates.}
    \begin{tabular}{c|c|c|c|c} 
        \hline
        Queries & Answer & CompGCN (Pre-trained) & tinyKG-BERT (Pre-trained) & VEM$^2$L (tinyKG-BERT) \\
        \hline
        (stapler, at location, ?) & desk & \makecell{1,(\underline{desk}, restaurant, winery, \\ street, classroom)} & \makecell{21,(town, shelf, any large city, \\market, drawer)} & \makecell{1,(\underline{desk}, town, school, \\shelf, drawer)} \\
        \hline
        (literature, at location, ?) & library & \makecell{5,(bookstore, university, \\ new york, theatre, \underline{library})} & \makecell{14,(university, class, \\ classroom, table, meet)} & \makecell{5,(university, theatre, \\ class, school, \underline{library})} \\
        \hline
        (sail, has property, ?) & fun & \makecell{1,(\underline{fun}, expensive, good \\ sport, cool, good)} & \makecell{140,(colorful, cool, good to \\ eat, warm, beautiful, \\ very tall)} & \makecell{2,(beautiful, \underline{fun}, cool, \\ colorful, expensive)} \\
        \hline
    \end{tabular}
\end{table*}


\subsection{Analyses}
\subsubsection{Hyperparameter $\mathcal{M}$ Analysis}
We observe that noisy neighbor triples affect performances of the fusion processes seriously, causing weak improvements on WN18RR discussed above. To explore the detailed influence of constructed neighbor triples, we conduct VEM$^2$L experiments with diverse $\mathcal{M}$ for graph densification. Results are presented in Table 3. As we can see, the changing process is roughly first rise and then fall, and the best point is at $\mathcal{M}=8$. Before the highest point, constructed neighbor triples further relieve the graph sparseness to some extent. While after that, more noises are generated and decrease the improvements.

Besides, $\mathcal{M}=0$ refers to ablating the graph densification module. Compared to $\mathcal{M}=8$, the ablation result attests to the importance of graph densification operation for relieving KG spareness.

\begin{figure}[t]
    \centering
    \includegraphics[width=0.93\linewidth]{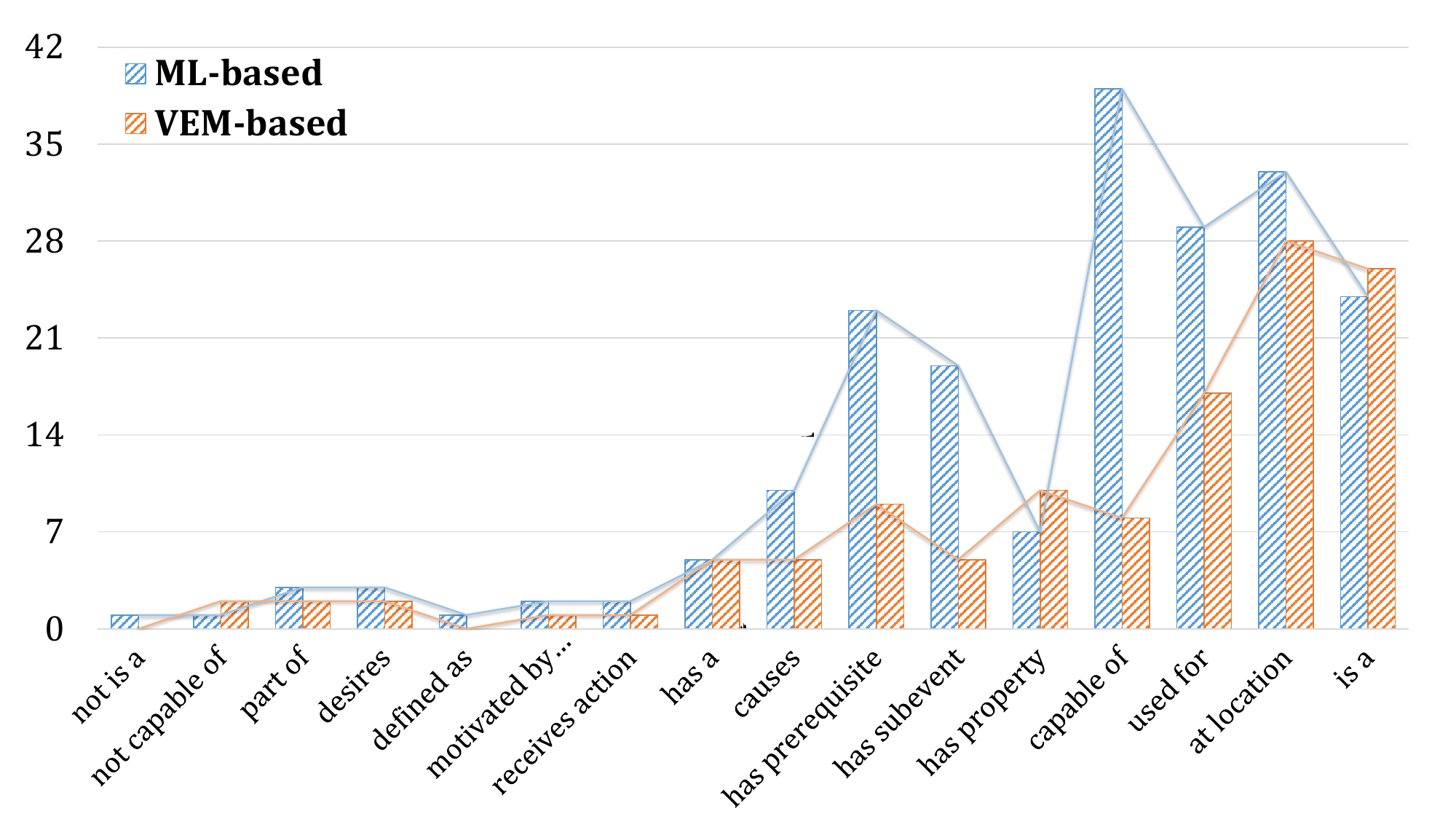}
    \caption{Number of improved queries on CN-100K for VEM-based and ML-based methods. x-axis is sorted according to the frequency of relations in training set.}
    \label{fig:conceptnet_vem_ml_improved}
\end{figure}

\subsubsection{Ablation Study}
The ML-based method fuses knowledge on training cases, thus we declare that this method focuses on expressiveness abilities. The VEM-based method fuses knowledge on constructed unobserved queries, so we think the VEM-based one can exchange generalization abilities. To explore the difference between ML-based and VEM-based methods, we dependently trained them separately and present quantitative results in Table 1. We observe that the ML-based method performs better than VEM-based one on the whole. This can be explained by that the VEM-based method trains upon constructed queries with more noises injected. We further analyze the improved queries to explore their detailed difference. Specifically, we count the number of improved queries after joint learning for each relation in Figure 5, note that we filter common improved cases to emphasize their respective improvements. 
It is clear that salient differences exist between these two methods, which may explain the phenomenon that using two methods together, i.e. VEM$^2$L, could achieve more improvements. 

\subsubsection{Implementation with other models.} To prove that our framework is agnostic to specific choices of structure-based and text-based models, we replace CompGCN and BERT with KBGAT and RoBERTa respectively, and experiment them on the CN-100K and FB15k-237\_20. According to the observation of \cite{sun2019re}, KBGAT suffers from the problem of abnormal score distributions. To this end, we modify the loss function from Sigmoid + BCELoss to Softmax + BCELoss. Due to the normalization constraint of Softmax function, the probability of different labels owning the same probability is very low. Experimental results presented in Table 5 show that our approach is universal.

\subsubsection{Case Study}
We have described the fusion process and operation mechanism in section 3. For better confirmation, we randomly list some queries in the CN-100K test where the queries experience a certain improvement after joint learning and present corresponding performances for structure-based and text-based models before and after fusing processes. As shown in Table 4, it is observed that pre-trained tinyKG-BERT underperforms on these queries compared with pre-trained CompGCN. After VEM$^2$L fusion processes, tinyKG-BERT achieves satisfactory ranking results with the help of high top-k recall performances offered by CompGCN. On the other hand, as demonstrated in the 1st and 2nd examples, we observe that tinyKG-BERT can properly retain acquired knowledge during pre-training while learning from CompGCN. Take the 1st query as an example, not only does tinyKG-BERT correct wrong ranks of ``dest", but retains other semantically-similar entities like ``shelf" and ``drawer" as top ranks, even if pre-trained CompGCN underestimates them. As a result, joint learning via the VEM$^2$L framework can drive the text-based model and structure-based model to learn from each other selectively, thus offering significant improvements compared to independent learning.

\begin{table}[t]
    \centering
    \caption{Different implementations to show the universality of VEM$^2$L. We replace CompGCN and BERT with KBGAT and RoBERTa respectively.}
    \begin{tabular}{cccccc}
    \toprule
    && \multicolumn{2}{c}{CN-100K} & \multicolumn{2}{c}{FB15k-237\_{20}}\\
    \cmidrule(lr){3-4}\cmidrule(lr){5-6}
    && Hits@10 & MRR & Hits@10 & MRR\\
    \midrule
    \multicolumn{2}{l}{BERT} & 0.588 & 0.367 & 0.286 & 0.174\\
    \multicolumn{2}{l}{KBGAT} & 0.465 & 0.267 & 0.305 & 0.179\\
    \multicolumn{2}{l}{VEM$^2$L} & 0.648 & 0.409 & 0.317 & 0.188\\
    \midrule
    \multicolumn{2}{l}{RoBERTa} & 0.574 & 0.363 & - & -\\
    \multicolumn{2}{l}{CompGCN} & 0.479 & 0.300 & - & -\\
    \multicolumn{2}{l}{VEM$^2$L} & 0.651 & 0.413 & - & -\\
    \bottomrule
    \end{tabular}
    \label{tab:table_2}
\end{table}

\section{Conclusion and Future Work}
In this paper, we study the task of sparse Knowledge Graph Completion. In order to ease the sparsity of KGs, we explore two improvements that utilize structure and text information and densify KGs dynamically. 
To fuse structure and text knowledge, we propose a novel plug-and-play fusion framework VEM$^2$L, consisting of two parts: ML-based and VEM-based methods, focusing on fusing the expressiveness ability and generalization knowledge respectively. We provide detailed proofs and exhaustive empirical analyses to verify the effectiveness of the proposed VEM$^2$L framework. Our future work intends to explore more stable graph densification methods and propose more effective methods to further utilize structural and textual knowledge. 

\bibliographystyle{ACM-Reference-Format}
\bibliography{reference}

\newpage

\appendix

\section{Proofs for Optimization Objectives}
In this appendix, we will provide proofs for the Theorem 1. and Theorem 2. given in VEM-based method respectively.

\subsubsection{Theorem 1.} \textit{Under the mean filed hypothesis, minimizing the KL divergence $D_{KL}(p_\pi({Y_U}|{Q_U})||p_\theta({Y_U}|{Y_L}, Q))$ equals to optimize the following objective:}
\begin{equation}
    \begin{aligned}
    &\max_\pi \quad \sum_{t_m \in \mathcal{T}_U}
    \{ \mathcal{H}(p_\pi(y_m|q_m)) \\ &+\sum_{y_m}\mathbb{E}_{p_\pi(Y_{\mathrm{NB}(t_m)\cap \mathcal{T}_U})}[\log p_\theta(y_m|q_m,\mathcal{T}_{\mathrm{NB}(t_m)})] \}
    \end{aligned}
    \nonumber
\end{equation}
\textit{Proof:} 
To make the notation more concise, we will omit $Q$ in the following proof. We first transfer the objective of KL divergence to the specific form:
\begin{equation}
    \begin{aligned}
    &D_{KL}(p_\pi(Y_U|Q_U)||p_\theta(Y_U|Y_L,Q))=\sum_{Y_U}p_\pi(Y_U)\log\frac{p_\pi(Y_U)}{p_\theta(Y_U|Y_L)}\\
    &=\sum_{Y_U}\prod_{t_n\in \mathcal{T}_U}p_\pi(y_n) \left[\log p_\pi(Y_U) - \log p_\theta(Y_L,Y_U) - \log p_\theta(Y_L) \right]\\
    &=\sum_{Y_U}\prod_{t_n\in \mathcal{T}_U}p_\pi(y_n) \left[ \sum_{t_n\in \mathcal{T}_U}\log p_\pi(y_n) - \log p_\theta(Y_L,Y_U) \right] + \text{const}\\
    \end{aligned}
    \nonumber
\end{equation}
Motivated by the Variational EM algorithm, if we consider each individual triple $t_{n_0}$, the objective function is given as follows:
\begin{equation}
    \begin{aligned}
    &\sum_{Y_U}\prod_{t_n\in \mathcal{T}_U}p_\pi(y_n) \left[ \sum_{t_n\in \mathcal{T}_U}\log p_\pi(y_n) - \log p_\theta(Y_L,Y_U) \right] + \text{const}\\
    &=\sum_{y_{n_0}}p_\pi(y_{n_0})\sum_{Y_{U\backslash n_0}}p_\pi(Y_{U\backslash n_0})[ \log p_\pi(y_{n_0})\\
    &\quad +\sum_{t_n\in \mathcal{T}_{U\backslash n_0}}\log p_\pi(y_n)
    - \log p_\theta(Y_L,Y_U) ] + \text{const}\\
    &=\sum_{y_{n_0}}p_\pi(y_{n_0})\log p_\pi(y_{n_0})+\sum_{y_{n_0}}p_\pi(y_{n_0})\sum_{Y_{U\backslash n_0}}p_\pi(Y_{U\backslash n_0})\\
    & \quad \log \prod_{t_n\in \mathcal{T}_{U\backslash n_0}}p_\pi(y_n) -\sum_{y_{n_0}}p_\pi(y_{n_0})\sum_{Y_{U\backslash n_0}}p_\pi(Y_{U\backslash n_0}) \\
    &\quad \log p_\theta(Y_L,Y_U) + \text{const}\\
    &=\sum_{y_{n_0}}p_\pi(y_{n_0})\log p_\pi(y_{n_0})-\sum_{y_{n_0}}p_\pi(y_{n_0})\sum_{Y_{U\backslash n_0}}p_\pi(Y_{U\backslash n_0})\\
    & \quad \log p_\theta(Y_L,Y_U) + \text{const}\\
    &=-\mathcal{H}(p_\pi(y_{n_0}))-\sum_{y_{n_0}}p_\pi(y_{n_0})\log F(y_{n_0}) + \text{const}\\
    &=-\mathcal{H}(p_\pi(y_{n_0}))-\mathbb{E}_{p_\pi(y_{n_0})} \left[ \log F(y_{n_0}) \right] + \text{const}  \label{16}
    \end{aligned}
    \nonumber
\end{equation}
Here, we introduce a variable $\log F(y_{n_0})$ for brevity:
\begin{equation}
    \log F(y_{n_0})=\sum_{Y_{U\backslash n_0}}p_\pi(Y_{U\backslash n_0})\log p_\theta(Y_L,Y_U)
    \nonumber
\end{equation}

\begin{table*}[t]
    \centering
    \caption{Detailed hyperparameters used in joint learning phases for VEMML framework}
    \begin{tabular}{c|ccccccccccccc}
    \toprule
    Datasets & \textit{tx\_vem\_s} & \textit{tx\_vem\_t} &\textit{ st\_vem\_s} & \textit{st\_vem\_t} & \textit{tx\_ml\_s} & \textit{tx\_ml\_t} & \textit{st\_ml\_s} & \textit{st\_ml\_t} & $\alpha_t$ & $\alpha_s$ & $\beta_t$ & $\beta_s$ \\
    \midrule
    FB15k-237\_20 & 1. & 1. & 1. & 1. & 1. & 1. & 1. & 1. & 1. & 1. & 6. & 4. \\
    WN18RR & 1. & 1. & 1. & 1. & 1. & 2. & 1. & 2. & 1. & 1. & 1. & 1. \\
    CN-100K & 1. & 5. & 1. & 1. & 1. & 5. & 1. & 1. & 1. & 1. & 4. & 1. \\
    \bottomrule
    \end{tabular}
    \label{tab:table_1}
\end{table*}

To further analysis $\log F(y_{n_0})$, we follow work of GMNN \cite{qu2019gmnn} to apply Markov Independence Hypothesis and have:
\begin{equation}
    \begin{aligned}
    \log F(y_{n_0}) &= \sum_{Y_{U\backslash n_0}}p_\pi(Y_{U\backslash n_0})\quad \log p_\theta(Y_L,Y_U)\\
    &=\mathbb{E}_{p_\pi(Y_{U\backslash n_0})}\left[ \log p_\theta(Y_L,Y_U) \right]\\
    &\approx \mathbb{E}_{p_\pi(Y_{U\backslash n_0})}\left[ \log p_\theta(y_{n_0}|Y_{\backslash y_{n_0}}) \right]\\
    &=\mathbb{E}_{p_\pi(y_{\mathrm{NB}(t_{n_0})\cap T_U})} \left[ \log p_\theta(y_{n_0}|Y_{\mathrm{NB}(t_{n_0})}) \right]
    \end{aligned}
    \nonumber
\end{equation}
Thus we can obtain equation:
\begin{equation}
    \begin{aligned}
    &D_{KL}(p_\pi(Y_U)||p_\theta(Y_U|Y_L)) = -\mathcal{H}(p_\pi(y_{n_0}))\\
    &\quad -\mathbb{E}_{p_\pi(y_{n_0})} \left[\mathbb{E}_{p_\pi(y_{\mathrm{NB}(t_{n_0})\cap T_U})} \left[ \log p_\theta(y_{n_0}|Y_{\mathrm{NB}(t_{n_0})}) \right] \right] + \text{const} \\
    &= -\mathcal{H}(p_\pi(y_{n_0})) -\sum_{y_{n_0}}\mathbb{E}_{p_\pi(y_{\mathrm{NB}(t_{n_0})\cap \mathcal{T}_U})} \left[ \log p_\theta(y_{n_0}|Y_{\mathrm{NB}(t_{n_0})}) \right] \\
    &\quad + \text{const}
    \nonumber
    \end{aligned}
\end{equation}
Considering each $t_m\in T_U$ and we get the final objective function:
\begin{equation}
    \begin{aligned}
    &\min \quad D_{KL}(p_\pi(Y_U)||p_\theta(Y_U|Y_L)) = \max \quad \sum_{t_m\in T_U} \{\mathcal{H}(p_\pi(y_{m}|q_m))\\
    &\quad +\sum_{y_{m}}\mathbb{E}_{p_\pi(y_{\mathrm{NB}(t_{m})\cap \mathcal{T}_U}|\mathrm{NB}(t_{m})\cap \mathcal{T}_U)} \left[ \log p_\theta(y_{m}|Y_{\mathrm{NB}(t_{m})}) \right]\}
    \nonumber
    \end{aligned}
\end{equation}



\subsubsection{Theorem 2.} \textit{Maximizing $\mathbb{E}_{p_\pi(Y_U|Q_U)}[\log p_\theta(Y_L,Y_U|Q)]$ equals to minimizing the following objective:}
\begin{equation}
    \begin{aligned}
    \mathcal{L}_{M}=&-\sum_{t_n\in \mathcal{T}_U}\mathbb{E}_{p_\pi(y_n|q_n)}[\log p_\theta(y_n|q_n, Y_{\mathrm{NB}(t_n)})]\\
    &-\sum_{t_m\in \mathcal{T}_L}\log p_\theta(\hat{y}_m|q_m, Y_{\mathrm{NB}(t_m) \cap \mathcal{T}_L})\\
    =&\mathcal{L}_{M,U}+\mathcal{L}_{M,L}
    \end{aligned}
    \nonumber
\end{equation}

\noindent \textit{Proof:} It is hard to maximize $\mathbb{E}_{p_\pi(Y_U|\mathcal{T}_U)}[log p_\theta(Y_L,Y_U|\mathcal{T})]$ directly. Motivated by GMNN \cite{qu2019gmnn}, we change to optimize the pseudolikelihood function as below:
\begin{equation}
    \begin{aligned}
    &\quad \mathbb{E}_{p_\pi(Y_U|\mathcal{T}_U)}[log p_\theta(Y_L,Y_U|\mathcal{T})]\\
    &\approx \mathbb{E}_{p_\pi(Y_U|\mathcal{T}_U)}[\sum_{t_n\in \mathcal{T}}\log p_\theta(y_n|t_n,Y_{U\backslash n})]\\
    &=\mathbb{E}_{p_\pi(Y_U|\mathcal{T}_U)}[\sum_{t_n\in \mathcal{T}}\log p_\theta(y_n|t_n,Y_{\mathrm{NB}(t_n)})]\\
    &=\mathbb{E}_{p_\pi(Y_U|\mathcal{T}_U)}[\sum_{t_n\in \mathcal{T}}\log p_\theta(y_n|t_n,Y_{\mathrm{NB}(t_n)\cap T_L})] \label{10}
    \end{aligned}
    \nonumber
\end{equation}

Above equation holds based on Markov Independence Hypothesis.
By dividing $\mathcal{T}$ into $\mathcal{T}_L$ and $\mathcal{T}_U$, we continue to obtain the following conclusion:
\begin{equation}
    \begin{aligned}
    &\mathbb{E}_{p_\pi(Y_U|\mathcal{T}_U)}[\sum_{t_n\in \mathcal{T}}\log p_\theta(y_n|t_n, \mathrm{Y}_{NB(t_n)\cap \mathcal{T}_L})]\\
    =&\mathbb{E}_{p_\pi(Y_U|\mathcal{T}_U)}[\sum_{t_n\in \mathcal{T}_U}\log p_\theta(y_n|t_n, \mathrm{Y}_{NB(t_n)\cap \mathcal{T}_L})\\ 
    &+ \sum_{t_m\in \mathcal{T}_L}\log p_\theta(\hat{y}_m|t_m, \mathrm{Y}_{NB(t_m)\cap \mathcal{T}_L})]\\
    \end{aligned}
    \nonumber
\end{equation}
Since we have hypothesized $p_\pi(Y_U|\mathcal{T}_U)=\prod_{t_n\in \mathcal{T}_U}p_\pi(y_n|t_n)$ by modeling $p_\pi$ with a textual encoding model, thus we have:
\begin{equation}
    \begin{aligned}
    &\mathbb{E}_{p_\pi(Y_U|\mathcal{T}_U)}[\sum_{t_n\in \mathcal{T}}\log p_\theta(y_n|t_n, \mathrm{Y}_{NB(t_n)\cap \mathcal{T}_L})]\\
    =&\mathbb{E}_{\prod_{t_n\in \mathcal{T}_U}p_\pi(y_n|t_n)}[\sum_{t_n\in \mathcal{T}_U}\log p_\theta(y_n|t_n, \mathrm{Y}_{NB(t_n)\cap \mathcal{T}_L})]\\
    &+ \sum_{t_m\in \mathcal{T}_L}\log p_\theta(\hat{y}_m|t_m, \mathrm{Y}_{NB(t_m)\cap \mathcal{T}_L})\\
    =&\sum_{t_n\in \mathcal{T}_U}\mathbb{E}_{p_\pi(y_n|t_n)}[\log p_\theta(y_n|t_n, \mathrm{Y}_{NB(t_n)\cap \mathcal{T}_L})]\\
    &+ \sum_{t_m\in \mathcal{T}_L}\log p_\theta(\hat{y}_m|t_m, \mathrm{Y}_{NB(t_m)\cap \mathcal{T}_L})
    \end{aligned}
    \nonumber
\end{equation}
where $\hat{y}_m$ is the ground-truth label for the labelled triple $t_m$. Therefore, maximizing $\mathbb{E}_{p_\pi(Y_U|\mathcal{T}_U)}[log p_\theta(Y_L,Y_U|\mathcal{T})]$ equals to minimize:
\begin{equation}
    \begin{aligned}
    \mathcal{L}_{M}=&-\sum_{t_n\in \mathcal{T}_U}\mathbb{E}_{p_\pi(y_n|t_n)}[\log p_\theta(y_n|t_m, Y_{\mathrm{NB}(t_n)})]\\
    &-\sum_{t_m\in \mathcal{T}_L}\log p_\theta(\hat{y}_m|t_m, Y_{\mathrm{NB}(t_m) \cap \mathcal{T}_L})\\
    =&\mathcal{L}_{M,U}+\mathcal{L}_{M,L}
    \end{aligned}
    \nonumber
\end{equation}



\begin{table}[h]
    \centering
    \caption{Explanations for temperatures}
    \begin{tabular}{c|ccc}
    \toprule
    Parameters & Method & Role & Target\\
    \midrule
    tx\_vem\_s & VEM-based & Student & $p_\pi$ \\
    tx\_vem\_t & VEM-based & Teacher & $p_\pi$ \\
    st\_vem\_s & VEM-based & Student & $p_\theta$ \\
    st\_vem\_t & VEM-based & Teacher & $p_\theta$ \\
    tx\_ml\_s & ML-based & Student & $p_\pi$ \\
    tx\_ml\_t & ML-based & Teacher & $p_\pi$ \\
    st\_ml\_s & ML-based & Student & $p_\theta$ \\
    st\_ml\_t & ML-based & Teacher & $p_\theta$ \\
    \bottomrule
    \end{tabular}
    \label{tab:table_7}
\end{table}

\section{experimental setups}
In this section, we produce more implementation details for the hyperparameter settings. Firstly, we introduce four trade-offs to weight importance of different losses, i.e. $\alpha_t$, $\alpha_s$, $\beta_t$, and $\beta_t$ in section 4.3.
Besides, following existing work of Knowledge Distillation, we also sped up fusion processes using temperature parameters. Specifically, we designed different temperature parameters for VEM-based and ML-based methods, which are explained in Table 7. For example, $tx\_vem\_s$ is the temperature used for logits outputted from the text-based model $p_\pi$ when $p_\pi$ acts as the student model during the VEM-based process. Detailed hyperparameters used in joint learning phases can be found in Table 6.


\end{document}